\documentclass[letterpaper, 10 pt, conference]{ieeeconf}  

\IEEEoverridecommandlockouts                              

\usepackage{times} 

\usepackage{graphicx}
\usepackage{amsmath,amssymb}
\usepackage{mathtools}
\usepackage{xargs}
\usepackage{bm}

\let\amssymbboxplus\boxplus
\let\amssymbboxminus\boxminus

\renewcommand{\boxplus}{\mathbin{\mathop\amssymbboxplus}}
\renewcommand{\boxminus}{\mathbin{\mathop\amssymbboxminus}}

\newcommand{ \xid        }{ \mathrm{x}            } 
\newcommand{ \yid        }{ \mathrm{y}            } 
\newcommand{ \zid        }{ \mathrm{z}            } 
\newcommand{ \tid        }{ \mathrm{t}            } 
\newcommand{ \effid      }{ \mathrm{e}            } 
\newcommand{ \coptrq     }{ \tau                  } 
\newcommand{ \efftrq     }{ \gamma                } 
\newcommand{ \effcross   }{ \kappa                } 
\newcommand{ \timeopt    }{ \Delta                } 
\newcommand{ \coms       }{ \mathbf{x}            } 
\newcommand{ \moms       }{ \mathbf{h}            } 
\newcommand{ \lmoms      }{ \mathbf{l}            } 
\newcommand{ \amoms      }{ \mathbf{k}            } 
\newcommand{ \forcevars  }{ \mathbf{F}            } 
\newcommand{ \lforcevars }{ \mathbf{\mathfrak{F}} } 
\newcommand{ \copvars    }{ \mathbf{\mathfrak{z}} } 
\newcommand{ \effpos     }{ \mathbf{p}            } 
\newcommand{ \effrot     }{ \mathbf{R}            } 
\newcommand{ \gravity    }{ \mathbf{g}            } 
\newcommand{ \robotmass  }{ m                     } 
\newcommand{ \setacteff  }{ \effid_{\mathrm{cnt}} } 
\newcommand{ \friccoeff  }{ \mu                   } 
\newcommand{ \dishorizon }{ N                     } 
\newcommand{ \fcost      }{ \phi                  } 
\newcommandx{\indexed}[5][1=,2=,4=,5=]{                                     
	\prescript{\mathrm{#1}}{\mathrm{#2}}{#3}^{\mathrm{#4}}_{\mathrm{#5}}
}

\usepackage{graphicx}

\usepackage[]{todonotes}

\usepackage{subcaption,caption} 

\usepackage{multirow} 
\usepackage{makecell} 

\newcommand{\norm}[1]{\left\lVert#1\right\rVert} 
\DeclareMathAlphabet{\pazocal}{OMS}{zplm}{m}{n}

\usepackage[noabbrev, capitalise]{cleveref}

\title{\LARGE \bf
Learning a Centroidal Motion Planner for Legged Locomotion
}

\author{Julian Viereck$^{1, 2}$, Ludovic Righetti $^{1, 2}$
\thanks{$^{1}$ Tandon School of Engineering, New York University, USA {\tt\small jviereck@nyu.edu, ludovic.righetti@nyu.edu}}%
\thanks{$^{2}$ Max Planck Institute for Intelligent Systems Tübingen, Germany}%
\thanks{This work was supported by the European Union’s Horizon 2020 research and innovation program (grant agreement 780684 and European Research Council’s grant 637935) and the National Science Foundation (grant 1825993).}%
}

\begin{document}

\maketitle
\thispagestyle{empty}
\pagestyle{empty}

\begin{abstract}

Whole-body optimizers have been successful at automatically computing complex dynamic locomotion behaviors. However they are often limited to offline planning as they are computationally too expensive to replan with a high frequency. Simpler models are then typically used for online replanning. In this paper we present a method to generate whole body movements in real-time for locomotion tasks. Our approach consists in learning a centroidal neural network that predicts the desired centroidal motion given the current state of the robot and a desired contact plan. The network is trained using an existing whole body motion optimizer. 
Our approach enables to learn with few training samples dynamic motions that can be used in a complete whole-body control framework at high frequency, which is usually not attainable with typical full-body optimizers.
We demonstrate our method to generate a rich set of walking and jumping motions on a real quadruped robot.

\end{abstract}

\section{Introduction}

Recently quadrupeds like Spot or AnyMAL have shown a new level of autonomy by traversing rough terrain. For control, these kind of robots often use a simplified dynamics model of the robot using only the dynamics of the center of mass~\cite{di2018dynamic}. In order to carry out more complex movements with a legged robot, it becomes important to take the full body dynamics into account. Recently, there has been progress in providing faster full body optimizers~\cite{tassa2012synthesis,mastalli2020crocoddyl,ponton2020efficient}. However, these methods can still require seconds to optimize a full-body movement over a sequence of contacts. 
This is too long to run the computation online while the robot is executing the motion. Reducing the optimization horizon can help decrease optimization time but to date, full nonlinear optimizer for legged robots are not run in fast control loops at the order of a few milliseconds.
Therefore, in many cases, either the full-body motion is computed up front and only replayed on the robot \cite{ponton2020efficient,carpentier2018multicontact,winkler2018gait} or a simplified model of the dynamics is used for online optimization \cite{di2018dynamic,daneshmand2020bipedal,mesesan2017dynamic}. This makes it difficult to react to (fast) changes in the environment that require full-body adaptation.

In this work, we present a new machine learning based approach for computing whole body movements for legged robots performing dynamic locomotion tasks. The method is fast and allows running the computation of full-body movements online at 100~Hz. Our approach uses the output of an existing kino-dynamic optimizer to train a planner 
capable of generating various motion patterns and generalize motions outside of the training data, e.g. to adapt the motion to new footstep  sequences. Furthermore, the resulting computation time is an order of magnitude faster than typical trajectory optimization methods.

\begin{figure}
  \centering
    \vspace{0.2cm}
    \includegraphics[width=1.0\columnwidth]{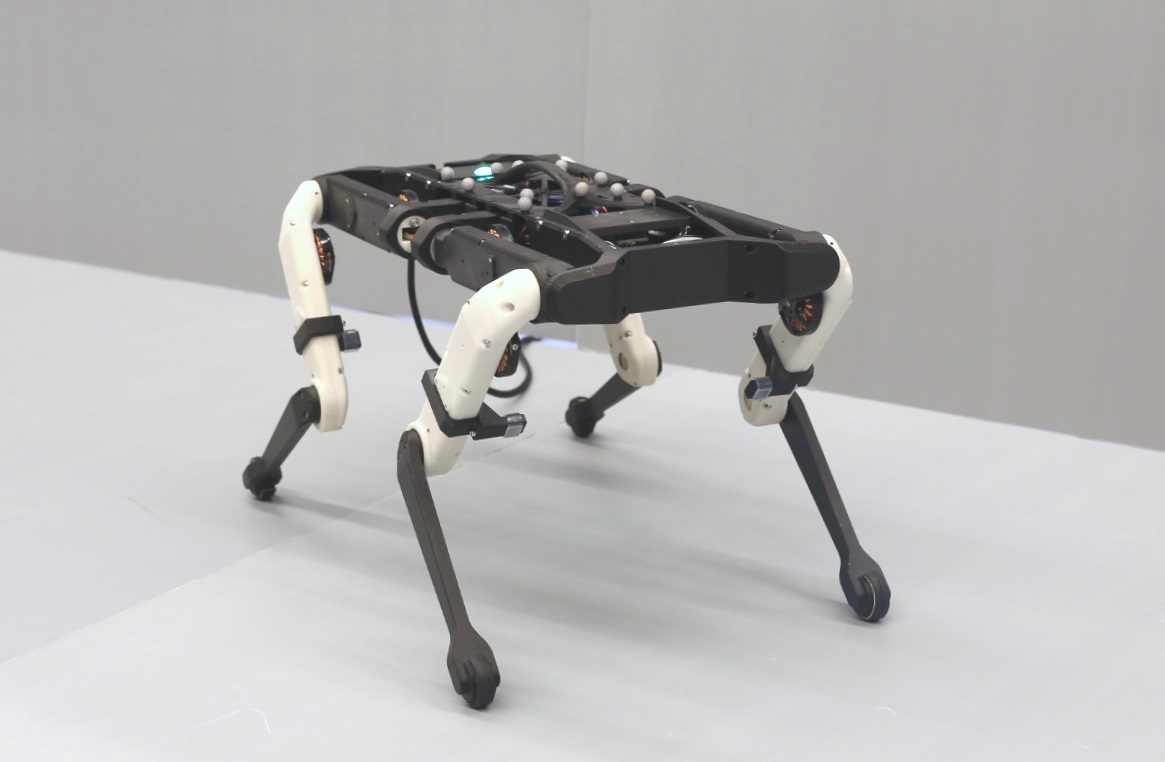}
    \setlength{\tabcolsep}{0.2em} 
    \setlength{\belowcaptionskip}{-10pt}
  \caption{The robot Solo12 used for the experiments \cite{grimminger2020open}.}
  \label{fig:robot_Solo12}
\end{figure}

The use of machine learning to cache the results of locomotion plans has been investigated before. In~\cite{mordatch2014combining} multiple tasks are optimized and a global neural network policy is trained from all the local optimal trajectories. In this work, however, we are not learning a policy but preserve a structured output from the motion optimizer.
\cite{melon2020reliable} uses a neural network to predict internal optimization variables to warm start a classical optimization method. While the method shows a speedup and improved initialization, it is not fast enough for real time use. Offline computations of full-body motions are also used to learn feasible contact transitions for efficient dynamically-consistent contact planning \cite{lin2020robust}.

The work closest to ours is the one presented in~\cite{kwon2020fast}. The approach also learns a network to predict the centroidal motions of a robot and intertwine the prediction with inverse kinematics. In contrast to our work, however, a full body motion generator is not used to generate the training data. Further, the input to the neural network is not relying on motion patterns as we do in this contribution. Finally, results, while impressive, are limited to simulations while we demonstrate our approach directly on a real quadruped.

Methods based on reinforcement learning have also gained popularity to compute full-body movements. They are model-free methods that typically sample from a simulator to learn policy. While these methods have shown  applications on real robots lately~\cite{peng2020learning, siekmann2020learning,bogdanovic2020learning}, their optimization is black-box and tend to require a lot of samples compared to the approach presented in this work.

The contributions of the paper are as follow: 1) we propose a method for learning and predicting the output of a whole body motion optimizer for a legged robot for dynamic tasks with contacts, 2) we introduce a way to learn motion patterns which allow to generalize the learned motions outside of the trained area, 3) we demonstrate the capabilities of our approach to generate motion plans at 100~Hz, one order of magnitude faster than the original motion planner, and 4) we demonstrate the applicability of our approach on walking and jumping experiments with a quadruped robot. 

\section{Background}
In this section, we describe the elements necessary for our approach, including the whole-body controller and inverse kinematics used to generate full movements and the kino-dynamic optimizer that will be replaced by our approach.

\subsection{Kino-dynamics motion optimizer}
We are interested in the kino-dynamic motion optimizer proposed in \cite{herzog2016structured}
that decomposes the problem into a centroidal dynamic optimization problem and a kinematic
problem. Given a motion description (i.e. sequence of contacts and cost function), the
solver alternatively optimize the dynamic and kinematic problems until they reach consensus.
While this method is very efficient \cite{ponton2020efficient} it still can take seconds to find a complete plan.

\subsubsection{Centroidal dynamic optimization}
The centroidal optimization problem, which our approach will aim to learn, is formulated as
\begin{subequations}
	\begin{align}
		\phantom{abcdefghi}
		\begin{aligned}
			%
			%
			\mathllap{\min_{\indexed{\moms}[][\tid], \indexed{\forcevars}[][\effid,\tid], \indexed{\copvars}[][\effid,\tid], \indexed{\coptrq}[][\effid,\tid]}}
			&\sum\limits_{\tid=1}^{\dishorizon} 
			%
			%
			 \indexed{\fcost}[dyn][\tid] \left(
				\indexed{\moms}[][\tid],
				\indexed{\copvars}[][\effid,\tid], \indexed{\forcevars}[][\effid,\tid], \indexed{\coptrq}[][\effid,\tid]
			\right) 
		\end{aligned}  \label{eq_dynopt_cost}
		\end{align}
\end{subequations}
subject to:
\begin{subequations}
\begin{align}
		&\begin{aligned}
			%
			%
			\moms_{\tid} =
			\begin{bmatrix}
				\indexed{\coms}[][\tid]  \\[0.5em]
				\indexed{\amoms}[][\tid] \\[0.5em]
				\indexed{\lmoms}[][\tid] \\[0.5em]
			\end{bmatrix} = 
			\begin{bmatrix}
				\indexed{\coms}[][\tid-1] + \frac{1}{\robotmass} \indexed{\lmoms}[][\tid] \indexed{\timeopt}[][\tid] \\[0.3em]
				\indexed{\amoms}[][\tid-1] + \sum\limits_{\effid \in \setacteff} \indexed{\effcross}[][\effid,\tid] \indexed{\timeopt}[][\tid] \\[0.6em]
				\indexed{\lmoms}[][\tid-1] + \robotmass \gravity \indexed{\timeopt}[][\tid] + \sum\limits_{\effid \in \setacteff} \indexed{\forcevars}[][\effid,\tid] \indexed{\timeopt}[][\tid] \\[0.0em]
			\end{bmatrix}
		\end{aligned} \hspace{-0.5cm} \label{eq_dynopt_momentum} \\
		&\begin{aligned}
			%
			%
			\mathllap{}
			&\; \indexed{\effcross}[][\effid,\tid] = (\indexed{\effpos}[][\effid,\tid] - \indexed{\coms}[][\tid]) \times \indexed{\forcevars}[][\effid,\tid] + \indexed{\efftrq}[][\effid,\tid]
		\end{aligned}  \label{eq_dynopt_kappa} \\
		&\begin{aligned}
			%
			%
			\mathllap{}
			&\; \indexed{\efftrq}[][\effid,\tid] = ( \indexed{\effrot}[\xid,\yid][\effid,\tid] \indexed{\copvars}[][\effid,\tid] )  \times \indexed{\forcevars}[][\effid,\tid] + \indexed{\effrot}[\zid][\effid,\tid] \indexed{\coptrq}[][\effid,\tid]
		\end{aligned} \label{eq_dynopt_gamma} \\
		&\begin{aligned}
			%
			%
			\mathllap{}
			&\; \indexed{\copvars}[\xid,\yid][\effid,\tid] \in [ \indexed[min]{\copvars}[\xid,\yid], \indexed[max]{\copvars}[\xid,\yid] ]
		\end{aligned} \label{eq_dynopt_cop} \\[0.0em]
		&\begin{aligned}
			%
			%
			\mathllap{}
			&\; \norm{ \indexed{\lforcevars}[\xid,\yid][\effid,\tid] }_{2} \le \friccoeff \indexed{\lforcevars}[\zid][\effid,\tid], \hspace{0.25cm} \indexed{\lforcevars}[\zid][\effid,\tid] > 0
		\end{aligned} \label{eq_dynopt_frccone} \\
		&\begin{aligned}
			%
			%
			\mathllap{}
			&\; \norm{\indexed{\effpos}[][\effid,\tid] - \coms_{\tid}}_{2} \le \indexed[max]{\pazocal{L}}[][\effid]
		\end{aligned} \label{eq_dynopt_eff_length} 
	\end{align}
	\label{dynopt_problem}
\end{subequations}
where the problem finds center of mass $\coms_{\tid}$, linear $\indexed{\lmoms}[][\tid]$ and angular $\indexed{\amoms}[][\tid]$ momentum, contact force $\indexed{\forcevars}[][\effid,\tid]$, center of pressure $\indexed{\copvars}[][\effid,\tid]$ and yaw torque $\indexed{\coptrq}[][\effid,\tid]$ at each contact to minimize 
user and consensus with the kinematics optimization costs $\indexed{\fcost}[dyn][\tid]$. 
$m$ is the mass of the robot, $\indexed{\effpos}[][\effid,\tid]$ the position of endeffector $\effid$ at time $\tid$, $\gravity$ is the gravity vector, $\indexed{\timeopt}[][\tid]$ the discretization time, $\indexed{\effrot}[][\effid,\tid]$ are rotation matrices describing the orientation of the endeffectors (z being the axis orthogonal to the surface), $ \friccoeff$ the friction coefficient, and $\indexed[max]{\pazocal{L}}[][\effid]$ the maximum distance between the CoM and an endeffector.
\cref{eq_dynopt_momentum} to \cref{eq_dynopt_gamma} ensure consistency with the centroidal dynamics, \cref{eq_dynopt_cop} are center of pressure bounds, \cref{eq_dynopt_frccone} are friction cone constraints and \cref{eq_dynopt_eff_length} ensures that the contact surfaces remain reachable. In this paper, we use the solver proposed in \cite{ponton2020efficient} to solve the problem.

\subsection{Inverse kinematics}
Given centroidal quantities and desired velocity of the endeffectors, we use a standard differential inverse kinematics algorithm for computing the corresponding whole body motion of the robot. We use three tasks: one tracking task on the centroidal motion which additionally stabilizes the base orientation, a task for each leg to track the desired endeffector velocity (zero velocity in case the endeffector should stay in contact with the ground) and a task to regularize the default posture of the robot. We use the pseudo inverse of the problem to compute joint and base velocities.
The computation along the trajectory is then as follows: We get a desired centroidal quantities for the current time step, we compute the inverse kinematics for the current time step, we integrate the velocity forward to obtain the new robot posture. Then the procedure repeats until the end of the trajectory.

\subsection{Whole body controller}
We use the whole body controller introduced in \cite{grimminger2020open}. The controller computes the desired wrench $\textbf{W}_\text{CoM}$\footnote{Note that the centroidal wrench is given by the centroidal force $\textbf{F}$ and centroidal momentum $\textbf{M}$ as $\textbf{W}_\text{CoM} = \begin{bmatrix}\textbf{F}\\\textbf{M}\end{bmatrix}$.} at the center of mass using a reference wrench $\textbf{W}^\text{ref}_\text{CoM}$ and a PD controller of the form
\begin{equation}
    \textbf{W}_\text{CoM} = \textbf{W}^\text{ref}_\text{CoM} + \begin{bmatrix}
        \textbf{K}_c (\indexed{\coms}[\text{ref}][] - \indexed{\coms}[][]) + 
        \textbf{D}_\text{c} (\indexed{\lmoms}[\text{ref}][] - \indexed{\lmoms}[][]) \\
        \textbf{K}_\textbf{b}(\textbf{q}^\text{ref}_\text{b} \boxminus \textbf{q}_\text{b}) +
        \textbf{D}_\text{b} ( \indexed{\amoms}[\text{ref}][] - \indexed{\amoms}[][])
    \end{bmatrix},
\end{equation}
where $\textbf{K}_c, \textbf{D}_\text{c}, \textbf{K}_\textbf{b}$ and $\textbf{D}_\text{b}$ are gains, $\indexed{\coms}[\text{ref}][]$ and $\indexed{\lmoms}[\text{ref}][]$, $\indexed{\amoms}[\text{ref}][]$ are the reference CoM position, linear and angular momentum, $\textbf{q}^\text{ref}_\text{b}$ is a quaternion for the desired base orientation.
$\indexed{\coms}[][]$ and $\indexed{\lmoms}[][]$, $\indexed{\amoms}[][]$ and $\textbf{q}_\text{b}$ are the corresponding measured quantities. The $\boxminus$ operator computes the difference between two quaternions as an angular velocity using the logarithmic map of SO(3).

To achieve the desired centroidal wrench, forces at the endeffectors in contact with the ground are allocated as 
\begin{align}
    \min_{\textbf{F}_i, \bm{\eta}, \zeta_1, \zeta_2} &\sum_{i} \textbf{F}^2_i + \alpha(\bm{\eta} + \zeta_1 + \zeta_2) \\
 \text{s.t.} ~~~ \textbf{W}_\text{CoM} = &\sum_{i \in C} \begin{pmatrix}\textbf{F}_i  \notag \\ \indexed{\effpos}[][i] \times \textbf{F}_i\end{pmatrix} + \bm{\eta} \\
 F_{i,x} < \mu F_{i,z} + \zeta_1, &F_{i,y} < \mu F_{i,z} + \zeta_2, 0 \leq F_{i,z}~~~\forall i \in C \notag,
\end{align}
where $C$ contains the indices of endeffectors currently in contact with the ground, $\textbf{F}_i$ is the desired force at the $i$th endeffector, $\indexed{\effpos}[][i]$ is the position of the $i$th endeffector with respect to the CoM, $\alpha$ is a large weight and $\bm{\eta}, \zeta_1, \zeta_2$ are slack variables. 

The torques for each leg $\bm{\tau}_i$ are computed using an impedance controller
\begin{align}
    \bm{\tau}_i = \textbf{J}^T_i \left(\textbf{F}_i + \textbf{K} (\textbf{I}^\text{ref}_i - \textbf{I}_i) + \textbf{D} (\dot{\textbf{I}}^\text{ref}_i - \dot{\textbf{I}}_i)\right),
\end{align}
where $\textbf{I}^\text{ref}_i$ and $\textbf{I}_i$ are the desired and measured endeffector positions.

\section{Learning a centroidal motion planner}
\begin{figure}
  \centering
    \vspace{0.2cm}
    \includegraphics[width=1\columnwidth]{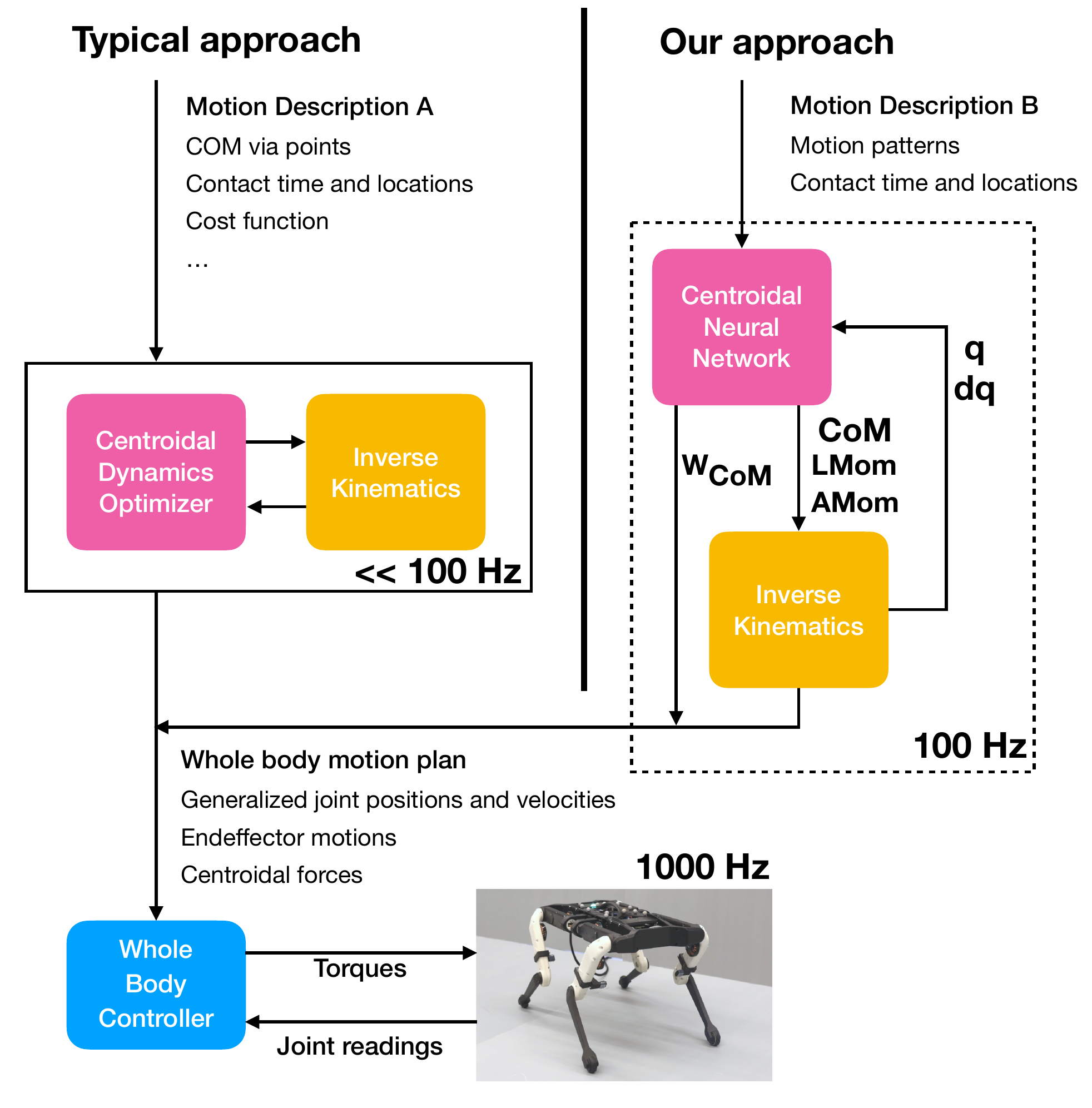}
    \setlength{\tabcolsep}{0.2em} 
  \caption{Pipeline overview: A motion description gets mapped to a whole body motion plan. In typical approaches (left side) this happens using a whole body motion optimizer, e.g. the kino-dynamic optimizer proposed in \cite{herzog2016structured}, which cannot be done in real time at 100 Hz. Our approach (right side): a centroidal neural network predicts the next desired centroidal quantities which are mapped to whole-body motions through inverse kinematics. The centroidal neural network gets feedback from robot state (positions and velocities). This method easily runs at 100 Hz. 
  The output of both methods is then sent to a whole-body torque controller (lower part).}
  \label{fig:overview_v3}
\end{figure}

%
%
In this section, we describe the propose approach to compute whole-body motions. \cref{fig:overview_v3} shows an overview of the approach.
Given a motion description (cost function, desired contact sequence and timing, etc), a motion planner computes the resulting whole-body motion plan. 
The plan contains full-body kinematic as well as dynamic trajectories, i.e. base position, joint positions, linear and angular momentum, contact forces, which is then tracked by a whole-body controller. 
A typical trajectory optimization approach would compute the whole body motion plan using an optimizer \cite{ponton2020efficient}. However, such optimizers are computationally expensive and cannot compute solutions at high frequencies (e.g. 100~Hz). 

In this paper, we propose to learn the motion optimizer instead to generate a typical gait, e.g. walking, jumping, etc, such that the computation can be done in real time. 
One straightforward manner to accomplish this would be to use motion plans computed by kino-dynamic optimizers to directly learn the entire whole-body motion plan at once. However, in practice such approach does not work very well as long-term predictions can be unstable and thereby make it impossible to predict a whole body motion plan over multiple time steps and contacts.

In contrast, we follow the (exact) kino-dynamic decomposition proposed in \cite{herzog2016structured,ponton2020efficient}. To compute the kinematic quantities of the whole body plan an inverse kinematics problem is solved tracking the previously computed centroidal quantities (center of mass, linear and angular momentum). Instead of computing the centroidal quantities, we predict these quantities using a previously trained neural network. We call this neural network the centroidal neural network, see~\cref{fig:overview_v3}. In this setting, the neural network has to predict the centroidal quantities for only a single time step into the future, instead of a full trajectory. In addition, the kinematic quantities like joint positions of the robot can be used as input to the network for prediction.

Using the centroidal network together with the inverse kinematics has also the benefit of using the ground truth dynamics model of the robot during the prediction. This helps to avoid diverging or unrealistic predictions.

\subsection{Motion patterns}

The motions we learn are often repetitive. For instance walking in a straight line is made up of the same motions at different offsets. Though these motions look similar, they would require to make different predictions for the centroidal network as the center of mass is moving over time. To avoid this, the centroidal network predictions are done in a local frame. For this, we divide the motion description into motion patterns. Each pattern specifies a local frame. 
In this local frame, the repeating motions look the same to the centroidal neural network and therefore it can predict the same output again.

\subsection{Centroidal neural network training and design}
\label{sec:centr_learning}

We use the motion optimizer in~\cite{ponton2020efficient} to compute  training data for the neural network. First, we optimize a motion description using the optimizer. Then, we record the inputs and the computed centroidal motions at every time step. Training the centroidal neural network is then a regression from the centroidal neural network's inputs to the computed centroidal motions.
We use the generalized positions and velocities of the robot from the last ten time steps as an input to the network. Here the base orientation is expressed in roll-pitch-yaw angles. We found it beneficial to add Gaussian noise on the generalized position and velocity training data. This helps to make the network prediction more robust to small divergences during the prediction process. 
Note that our centroidal neural network is not using all the information from the motion description. For instance the information about CoM via points is not used. The network still learns how to predict the motion patterns properly even when using less information.

In addition to the generalized positions and velocities, we also use contact information from the last 50 until 50 future time steps as input. We found it necessary to have a very large range of contact information provided to the neural network for predicting long ranging effects like swinging the body back after landing. The contact information comes from the motion description (e.g. they can be computed with a contact planner \cite{tonneau2018efficient}). At every time step each endeffector is either in contact with the ground at a specified position or moving to the next desired contact location. If the endeffector is planned to be in contact with the ground we put a contact duration of zero. Otherwise we encode the time till making the next contact. The contact information for each endeffector at each time step is then the contact location and contact duration. All quantities are expressed in the local frame of the current motion pattern.

In our evaluations, we noticed that the last motion pattern before the robot stops behaved sometimes differently. This is due to the motion optimizer trying to minimize the linear and angular momentum towards the end. In this case, the centroidal network has to change its prediction when the robot needs to stop. To indicate if a motion pattern is the last one to the centroidal network, a binary flag is added to the input. The binary flag is set to ten if the current motion pattern is the last one and otherwise the value is set to zero.

While in principle the centroidal force and centroidal momentum can be computed by taking the derivative of the linear and angular momentum, we found these differentiated quantities to be too noisy to use on the real robot. Therefore, we also added these quantities as outputs of the centroidal neural network. Similarly, the CoM position can be computed by integrating the linear momentum. We found the resulting CoM trajectory to drift away from the desired one. Therefore, we are predicting the CoM position from the centroidal neural network and fuse it with the linear momentum prediction in the inverse kinematics step.

The centroidal neural network is modeled as a feedforward neural network. The first hidden layer is made up of 32 neurons and uses a soft sign activation function. Afterwards there are two more hidden and one output layer using soft sign for the first layer and ReLU activation functions for the second layer. Each layer comes with 128 hidden neurons. During experiments we found it crucial for the prediction performance to have the initial bottleneck with 32 neurons and a saturating output function like soft sign. The output is of size 15: nine outputs represent the three dimensions (x, y, z) each for the center of mass, linear and angular momentum; six outputs represent the centroidal wrench. The network regression is optimized using Adam optimizer using a learning rate of 1e-4 and weight decay of 1e-4. The batch size is 256 samples and we optimize for 64 epochs (less than one minute on a GPU). The regression is done using a L1 loss between the center of mass, linear, angular momentum and centroidal wrench separately.

\section{Experiments}

\begin{figure*}
    \centering
    \includegraphics[width=\textwidth]{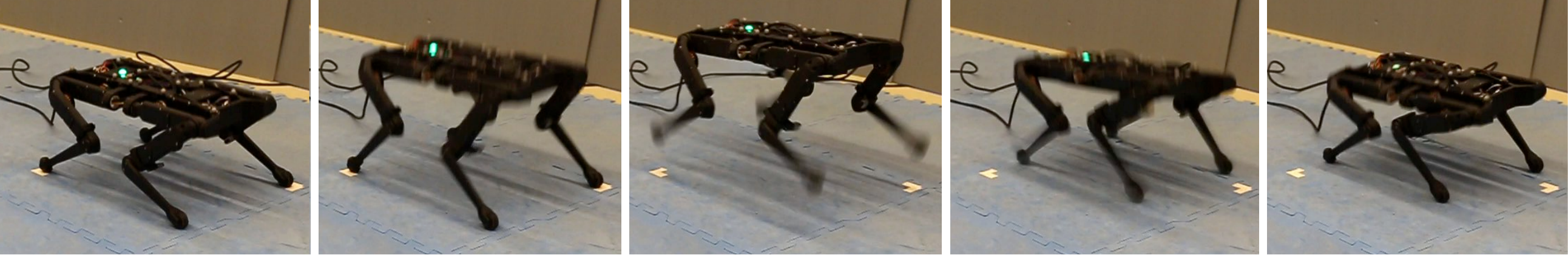}
    \vspace{-0.3cm}
    \setlength{\belowcaptionskip}{-10pt}
    \caption{Screenshots from a sideways jumping motion. The images are 0.1 s apart.}
\end{figure*}

All experiments are performed on a real Solo12 quadruped robot (\cref{fig:robot_Solo12}), an open-source \cite{ODRI, grimminger2020open} torque-controller robot with 12 actuated degrees of freedom (shoulder, hip and knee joints for each leg). In our experiments the absolute base position and orientation of the robot were measured using a motion capture system.

To assess the quality of the generated motions on the robot, 
we compare the centroidal network predictions compared to the computed motions with the kino-dynamics optimizer. In particular, we look at the tracking performance using the whole-body controller to assess the dynamic feasibility of the generated motions.
We compute the tracking error between the planned trajectory and the tracked trajectory when executing the original plan and the plan generated by the centroidal network. The tracking error is computed as the distance error for the center of mass (COM) and the base orientation. We report the mean tracking error as well as the maximum tracking error.

In the following we demonstrate our method on three example motions: first a set of static walks that we compare with the kino-dynamic optimizer, then a longer sequence of static walks that were not part of the training data, and finally a set of jumping motions. We answer the following questions with our experiments: 1) Is our method able to make predictions in real time, 2) is our method able to learn motion patterns that can be executed on a robot and 3) is our method able to generate movements with variables desired contact sequences?

\subsection{Static walk motion}
We create a motion generator that allows us to step in any direction. The motion generator performs always the same sequence of leg motions: front left, hinge right, front right, hinge left. For training our centroidal neural network, we generate motions with three consecutive static walks using our motion generator. This results in a walk of 2.9 seconds duration. Each of the walks is in a randomly sampled direction. We assign a different motion pattern to each of these static walks. We generate 60 motion descriptions and use them to train the centroidal neural network described in~\cref{sec:centr_learning}.

To test the quality of the learned centroidal neural network, we generate eight test static walks samples with ten consecutive static walks each with a total duration of 14.1 seconds. As with the 60 training samples the directions of the static walks are sampled randomly.

\begin{figure*}
    \centering
    \begin{subfigure}{0.45\textwidth}
    \includegraphics[width=\columnwidth]{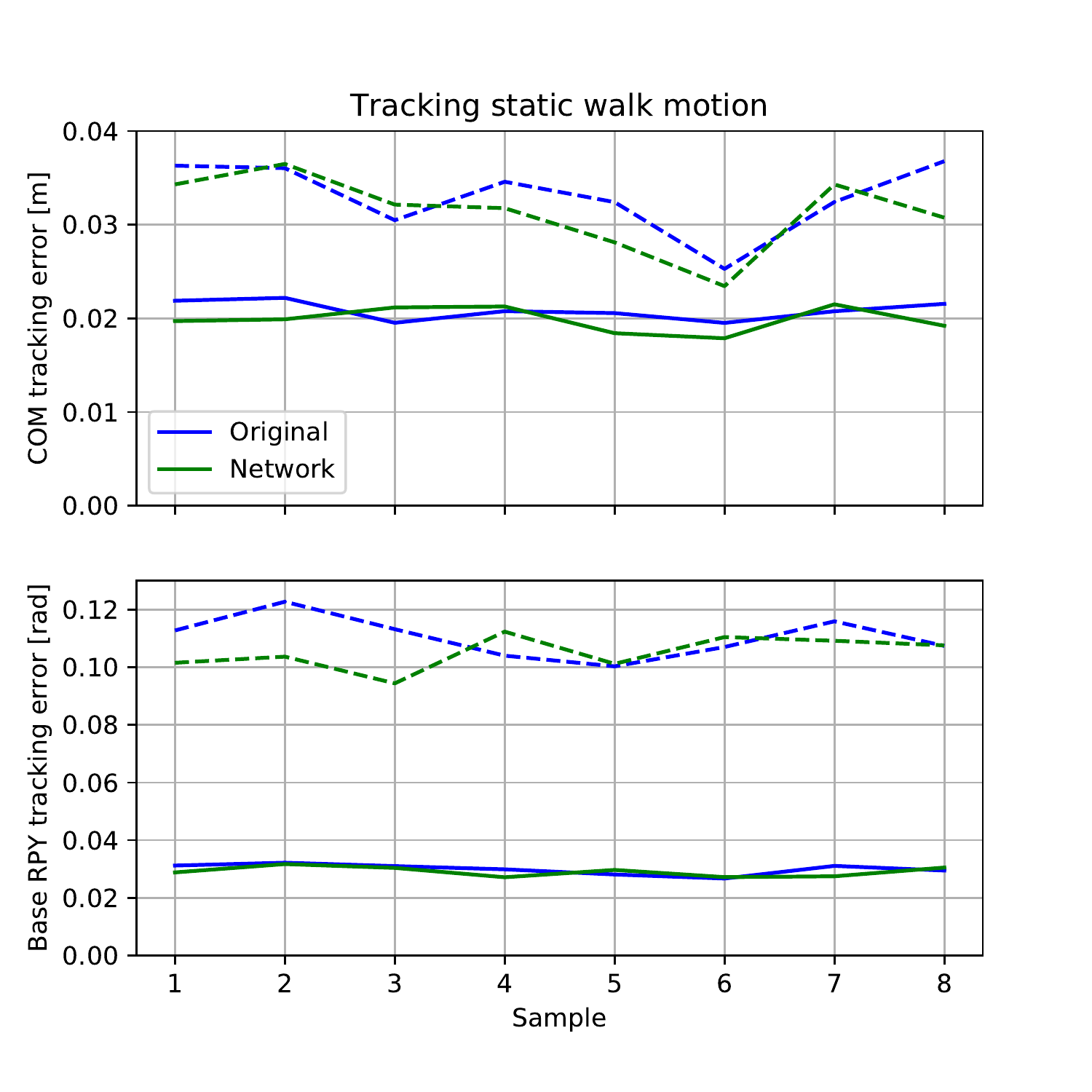}
    \vspace{-0.8cm}
    \caption{Tracking error during walking}\label{fig:result_static}
    \end{subfigure}
    \begin{subfigure}{0.45\textwidth}
    \includegraphics[width=\columnwidth]{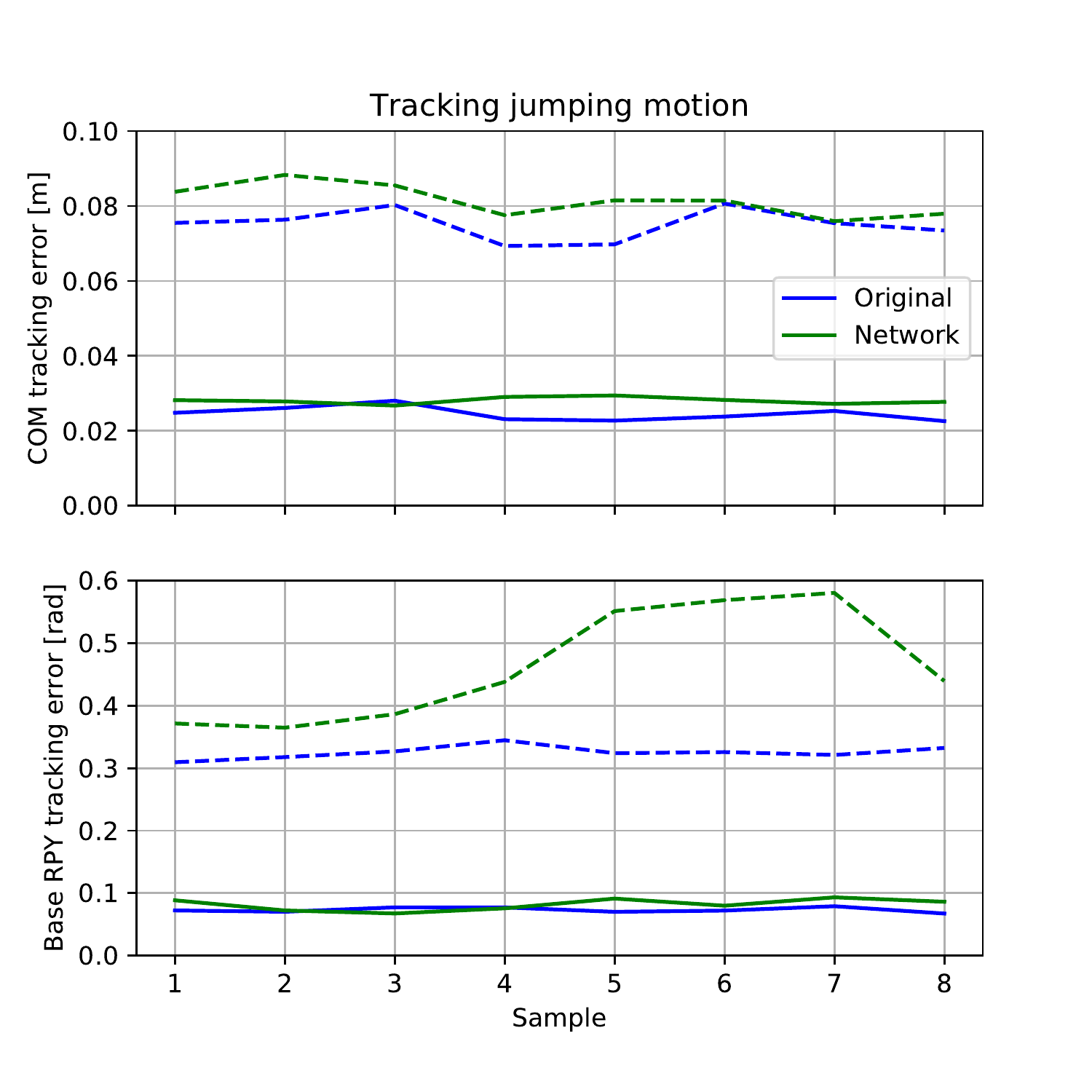}
    \vspace{-0.8cm}
    \caption{Tracking error during jumping}
    \label{fig:result_jumping}
    \end{subfigure}
        \caption{(Left) Results from tracking eight random static walks over ten steps on the robot. (Right) Results from tracking  eight   motions   with   eight random  jumps  each on the robot. In blue we show results with the original plan optimized with the kino-dynamic optimizer~\cite{ponton2020efficient} (labeled "Original") and in green with the plan generated using the centroidal neural network (labeled "Network"). Top plot shows the tracking error for the CoM; lower plot shows the tracking error for the base angular orientation. Solid line is the mean tracking error while dash line shows the maximum tracking error. The centroidal network is capable of producing motions of similar quality as the original planner.}
\end{figure*}

The tracking results for the static walks are shown in~\cref{fig:result_static}. From the plots we see that the tracking of the center of mass as well as the base orientation is very similar between the original plan and the network generated plan. Overall, the robot is able to track all motions generated by the neural network without falling.

\subsection{Marathon motion}
This motion tests the ability of the method to generate long lasting motions. We randomly generate a stepping sequence made up of 50 steps and test the ability of our approach to generate motions for long step sequences. The results of the marathon task are shown in the video provided with this submission. As one can see, the robot is able to execute the 50 static walk steps without any problem, although the sequence of steps and associated motion plans were not part of the training data.

\subsection{Jumping motion}
Beside walking, we are interested in more dynamic motions, such as jumping. We create a motion generator that generate jumps in random directions. Similar as before for the walking motions, we generate 60 motion descriptions consisting of three different jumps as training data. Each jump is assigned a different motion pattern.
For testing the learned centroidal network, we generate eight jumping motions with 10 jumps each. Each jump goes into a randomly selected direction, to demonstrate the  ability of the approach to generate movements from previously unseen step sequences.

The results for the jumping task are shown in~\cref{fig:result_jumping}. As one can see, the COM tracking is very similar when tracking the original plan and when tracking the plan generated by the neural network. For the tracking of the base orientation, the mean tracking error is again similar between the two tracking methods. However, the maximum tracking error when using the neural network plan is up to nearly twice as large as the original plan in three out of the eight cases. Again, the robot managed to track all plans generated using the centroidal neural network without falling.

\begin{figure}
    \centering
    \includegraphics[width=\columnwidth]{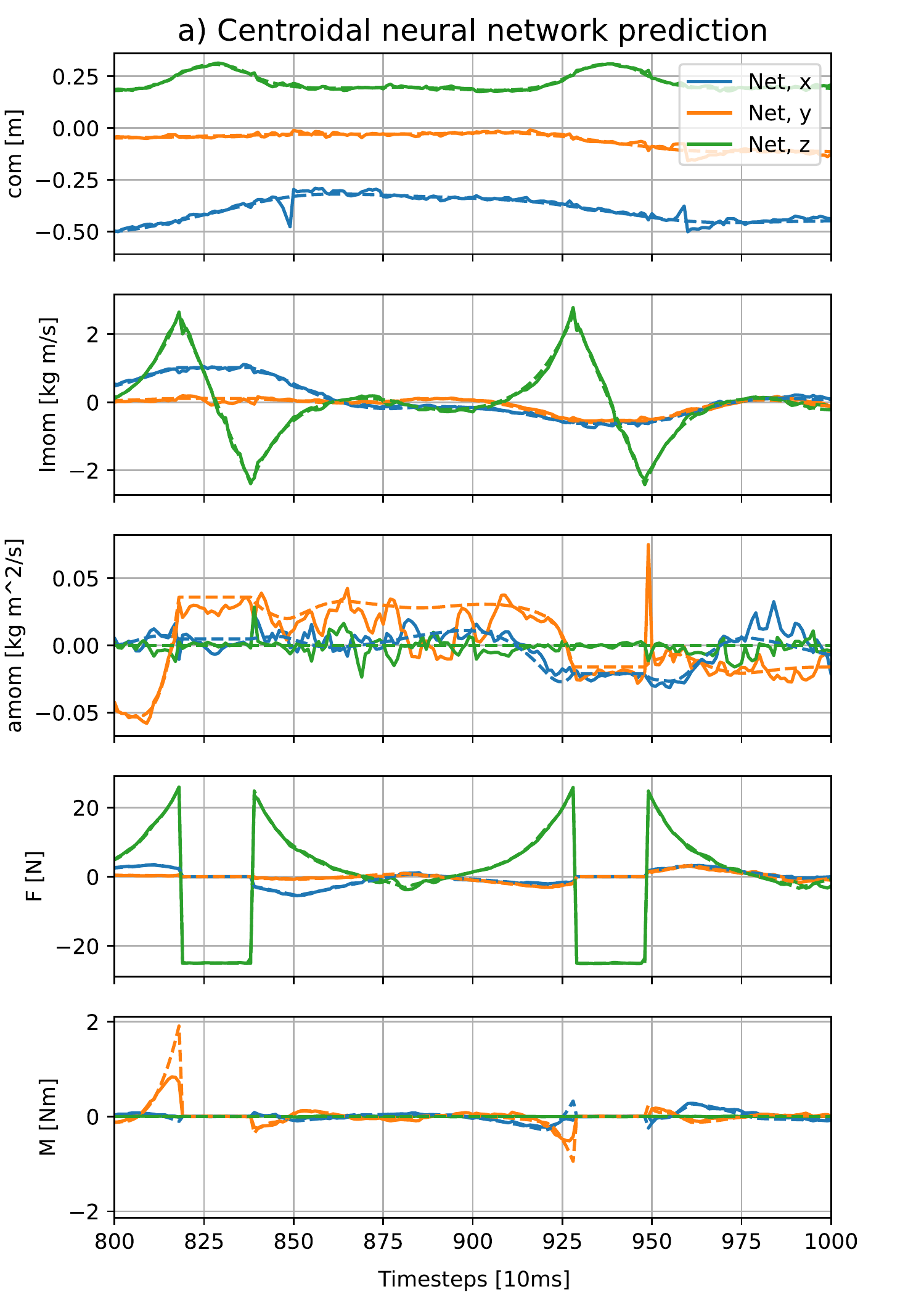}
    \vspace{-0.5cm}
    \setlength{\belowcaptionskip}{-10pt}
    \caption{Comparing the centroidal neural network predictions to the original plan predictions for two seconds along one of the test jump sample trajectories. The output from the centroidal neural network is plotted as solid lines, the original plan as dashed lines.}
    \label{fig:prediction}
\end{figure}

\begin{figure}
    \centering
    \includegraphics[width=\columnwidth]{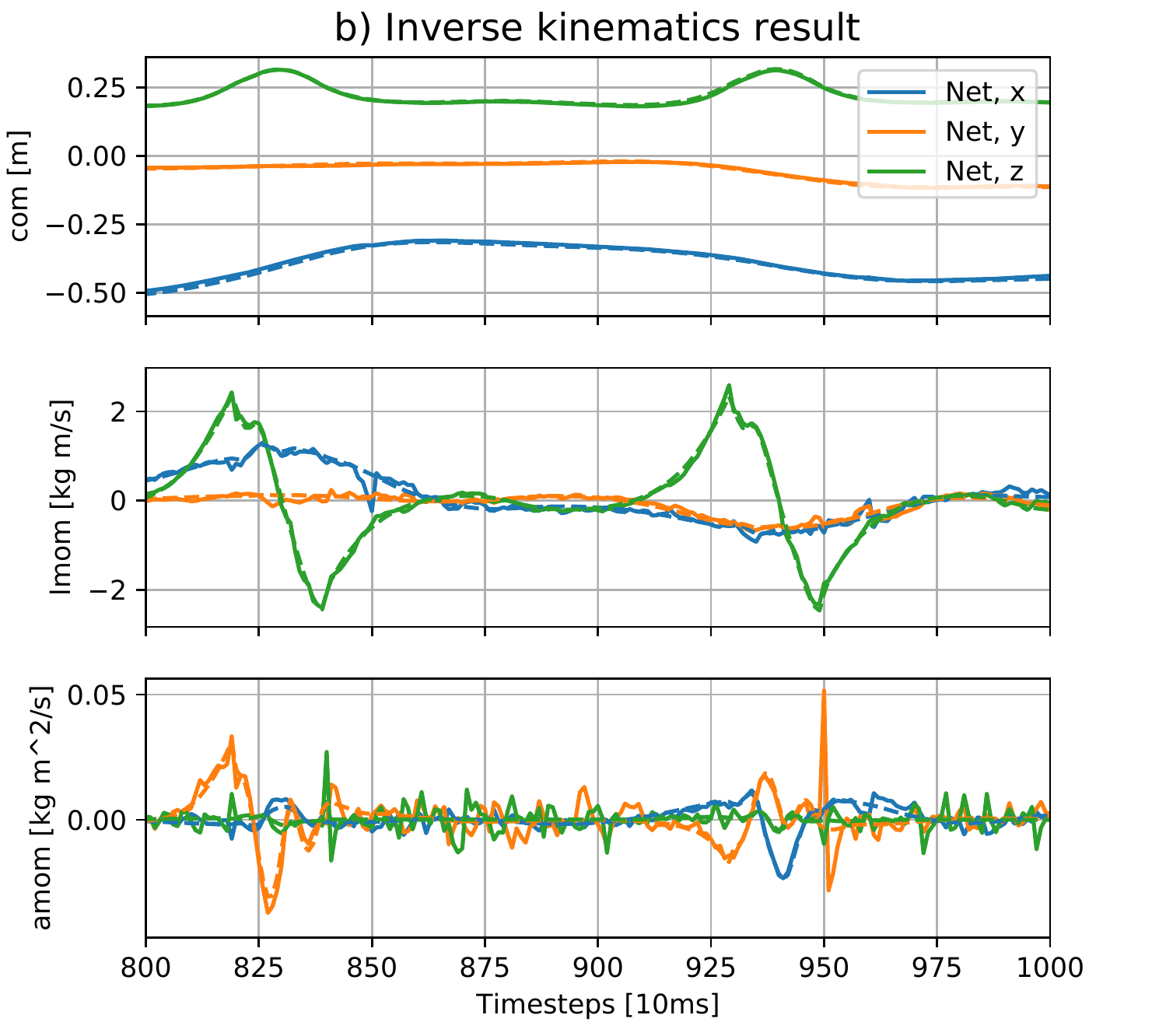}

    \caption{Comparing result of running inverse kinematics based on the centroidal neural network input from~\cref{fig:prediction} and the original plan for two seconds along one of the test jump sample trajectories. The output from centroidal neural network inverse kinematics is plotted as solid lines, the original plan as dashed lines.}
    \label{fig:prediction_kinematics}
\end{figure}

For one of the jumping task, detailed plots of the centroidal neural network prediction are shown in~\cref{fig:prediction}. We notice that it can generate predictions very close to what the original planner would compute (i.e. close to the optimal motions).
The result after running the inverse kinematics for the same section is shown in~\cref{fig:prediction_kinematics}. We notice that the network generates kinematically feasible motions. Further, the inverse kinematic part plays an important role as it filters the motion before sending it to the whole-body controller.

\subsection{Timing}

Timing information about the motions and how long it takes to optimize them using the original method and the centroidal neural network is shown in~\cref{table:timing}. Here we show the time it takes for the network to generate a complete plan when integrating forward for the whole duration of the motion. As one can see, the neural network method is significant faster: For the walking motion it is 22.1 times faster than the original method and up to 41.3 times faster on the jumping motion. Note in addition, that the prediction time is always lower than the plan duration for the network method. This makes it possible to predict the plan in real time.

{\setlength{\tabcolsep}{0.5em} 
\renewcommand{\arraystretch}{1.2}
\begin{table}[h]
\begin{center}
\begin{tabular}{|c|c|c|c|c|}
\hline
\multirow{2}{*}{Task} & \multirow{2}{*}{Motion duration} & \multicolumn{2}{c|}{Method computation time} & \multirow{2}{*}{Speedup} \\
& & ~~Original~~ & Network & \\
\hline
Walk & 14.1 s & 35.4 s & 1.6 s & 22.1$\times$ \\
\hline
Marathon & 70.1 s & 236.6 s & 8.3 s & 28.5$\times$ \\
\hline
Jump & 11.0 s & 99.1 s & 2.4 s & 41.3$\times$ \\
\hline
\end{tabular}

\end{center}
\setlength{\belowcaptionskip}{-10pt}
\caption{Timing information for the different motions. The table shows how long each motion is and how long it takes to compute the plan for the motion using the original kino-dynamics optimizer~\cite{ponton2020efficient} and the proposed neural network method. The speedup is how much faster the centroidal neural network method is over the original method.}
\label{table:timing}
\end{table}}

\subsection{Discussion}

During the motions the robot moves outside of the area it was trained on. This shows that using the motion patterns the robot is able to extrapolate motions to new areas for the same repetitive motions. In addition, as seen in~\cref{table:timing} our method is able to make predictions in real time as the compute time is lower than the plan duration for all the motions. Last but not least, the proposed method was able to generate plans for static walks as well as for jumping motions. This demonstrates its ability to generate motions for dynamic tasks involving multiple switching contacts.

\section{Conclusion and Future Work}

In this paper, we considered the problem of speeding up a whole body motion planner to real time. We do this by learning a centroidal neural network. We use the centroidal neural network together with an inverse kinematics solver to solve for a whole body plan, which is then tracked using a whole body controller. In our experiments we can show that we achieve this goal and predict faster than real time. The presented method has been validated for three tasks that are about static walking in different directions as well as jumping in different directions. This demonstrates that the method works for contact rich, dynamic tasks. The use of motion patterns allows us to generalize to new areas outside of the training data.
While the current motions have only been evaluated on flat ground thus far, we are planning in future work to extend the motions to more complex non-planar scenarios and to study extensions of the approach to more complex robots such as a humanoid.

\addtolength{\textheight}{-12cm}   




\bibliographystyle{IEEEtran}
\bibliography{IEEEabrv,references}

\end{document}